\documentclass[10pt,twocolumn,letterpaper]{article}

\usepackage[pagenumbers]{cvpr} 
\usepackage{multirow}
\usepackage{xcolor}
\definecolor{lightblue}{RGB}{230,245,255}

\usepackage{float}
\usepackage{placeins}

\definecolor{cvprblue}{rgb}{0.21,0.49,0.74}
\usepackage[pagebackref,breaklinks,colorlinks,allcolors=cvprblue]{hyperref}
\usepackage{cuted}
\usepackage{tabularx}
\usepackage{pifont}

\usepackage[table]{xcolor}
\definecolor{best}{RGB}{255, 179, 179}  
\definecolor{second}{RGB}{255, 204, 153}
\definecolor{third}{RGB}{255, 255, 153}  

\newcommand{\best}[1]{\cellcolor{best}\textbf{#1}}
\newcommand{\second}[1]{\cellcolor{second}#1}
\newcommand{\third}[1]{\cellcolor{third}#1}
\def\paperID{14344} 
\def\confName{CVPR}
\def\confYear{2026}

\newcommand{\ours}{\texttt{Flow3r}}
\newcommand{\oursvggt}{\texttt{Flow3r-VGGT}}

\newcommand{\expa}{\small{\texttt{3d-sup}}}
\newcommand{\expb}{\small{\texttt{flow-projective}}}
\newcommand{\expc}{\small{\texttt{flow-tracking}}}

\newcommand{\expd}{\texttt{flow-factored}}

\title{Flow3r: Factored Flow Prediction for Scalable Visual Geometry Learning}

\author{
Zhongxiao Cong \quad Qitao Zhao \quad Minsik Jeon \quad Shubham Tulsiani \\
Carnegie Mellon University \\
{\small \textit{project \& code}: \textcolor{blue}{https://flow3r-project.github.io/}}}

\begin{document}
\maketitle

\begin{strip}
\vspace{-40pt}
\centering
\includegraphics[width=1.0\textwidth]{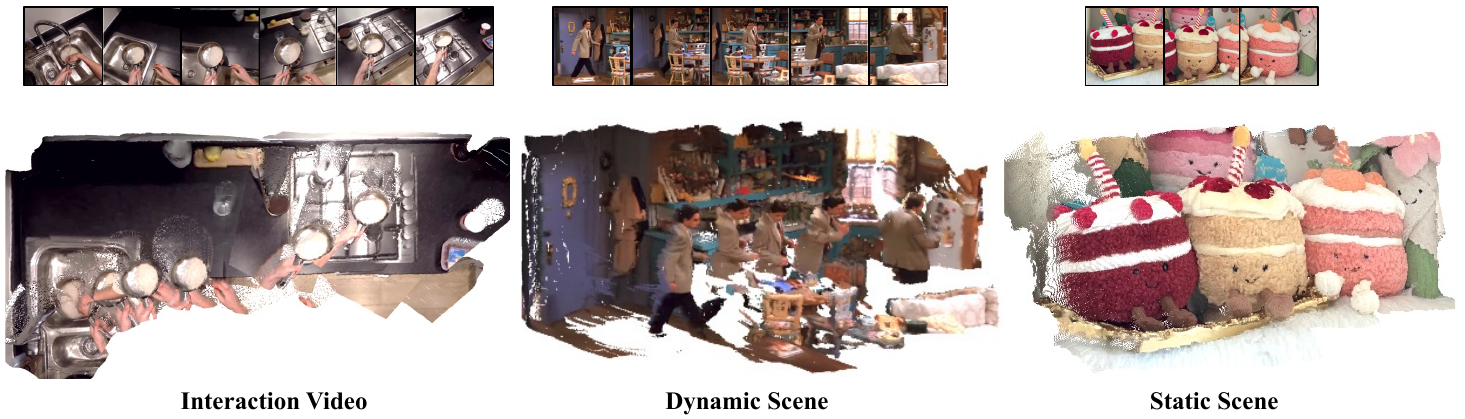}
\captionsetup{hypcap=false}\captionof{figure}{
\ours{} leverages \emph{unlabeled} videos (using flow supervision) alongside labeled 3D data for scalable visual geometry learning. This enables accurate multi-view 3D reconstruction in-the-wild, in particular for settings with scarce labeled data \eg, interaction videos and dynamic scenes.
}
\vspace{-0.1in}
\label{fig:teaser}
\end{strip}

\vspace{6pt}
\begin{abstract}
Current feed-forward 3D/4D reconstruction systems rely on dense geometry and pose supervision -- expensive to obtain at scale and particularly scarce for dynamic real-world scenes. We present \ours, a framework that augments visual geometry learning with dense 2D correspondences (`flow') as supervision, enabling scalable training from unlabeled monocular videos. Our key insight is that the flow prediction module should be \emph{factored}: predicting flow between two images using geometry latents from one and pose latents from the other. This factorization directly guides the learning of both scene geometry and camera motion, and naturally extends to dynamic scenes.
In controlled experiments, we show that factored flow prediction outperforms alternative designs and that performance scales consistently with unlabeled data.
Integrating factored flow into existing visual geometry architectures and training with ${\sim}800$K unlabeled videos, \ours{} achieves state-of-the-art results across eight benchmarks spanning static and dynamic scenes, with its largest gains on in-the-wild dynamic videos where labeled data is most scarce.
\end{abstract}    
\section{Introduction}
\label{sec:intro}

The task of `visual geometry inference', \ie, recovering the 3D structure of a (static or dynamic) scene from multi-view input images has undergone a  paradigm shift -- evolving from classical optimization-based methods~\cite{agarwal2011building,frahm2010building,liu2024robust,schonberger2016structure,snavely2006photo,wu2013towards} to recent data-driven predictors~\cite{wang2023DUSt3R,wang2025vggt,zhao2025diffusionsfm, wang2025continuous} that can directly output the geometry and pose corresponding to the input images. The success of such systems, however, has crucially relied on multi-view training data with dense geometry and camera pose supervision.
Unfortunately, such supervision is not easily available across all settings of interest, \eg, for dynamic scenes in the wild or domains like egocentric videos, and existing methods do not generalize well to such scenarios. More broadly, unlike the self-supervised objectives that have enabled scaling of LLMs and vision transformers, the reliance on dense geometry and pose labels prevents truly large-scale visual geometry learning.

We take a step towards scalable learning of multi-view models and present \ours, a framework to guide visual geometry learning from \emph{unlabeled} videos, \ie, without any explicit geometry or pose supervision. Instead, \ours~leverages a readily available supervisory signal that is a cornerstone of classical (and recent) optimization-based multi-view methods -- dense \emph{correspondences across images}. Inspired by recent progress in inferring dense correspondences for generic image pairs and videos, we seek to unlock scalable learning by incorporating such 2D flow as auxiliary supervision for 3D visual geometry models. The key technical question we address is: \textbf{\emph{`how can flow
effectively supervise visual geometry prediction?'}}.

We are not the first to consider flow supervision for visual geometry learning. Indeed, VGGT~\cite{wang2025vggt} adds a `tracking' module that uses local features from two images to predict flow between them as an auxiliary training objective. However, as we show later, this merely encourages visually discriminative features but does not directly aid the learning of pose or geometry. Our key insight is that, to guide geometry learning, the \emph{{flow prediction module should be asymmetric}}. We observe that for static scenes, the flow between a source and a target image can be induced only via the geometry of the source image (pointmaps in a global coordinate) and the camera pose of the target. Motivated by this, we propose to incorporate \mbox{a~ \emph{\textbf{factored flow prediction module}}} in visual geometry models. Specifically, such models typically compute `local' patchwise features that later predict geometry as well as a global per-image token that infers camera pose. Our flow prediction module computes flow between two images using \emph{{only the global pose token for the target along with the patchwise tokens for the source}}.

We find that factored flow prediction provides stronger supervision for pose and geometry than the `tracking' design adopted by previous works. Moreover, unlike projection-based flow inference, it enables robust prediction and naturally extends to dynamic scenes. \ours~integrates factored flow prediction and leverages ${\sim}800$K unlabeled videos as supervision alongside existing labeled 3D datasets. This allows \ours~to achieve state-of-the-art results across diverse benchmarks spanning static and dynamic scenes, with the largest improvements on in-the-wild dynamic videos where labeled data is scarce. More broadly, by extracting the supervisory signal from unlabeled videos, \ours{} represents a step towards large-scale visual geometry learning without large-scale labeled data.

\section{Related Work}
\label{sec:related}

\begin{figure*}[t]
  \centering
  \includegraphics[width=\linewidth]{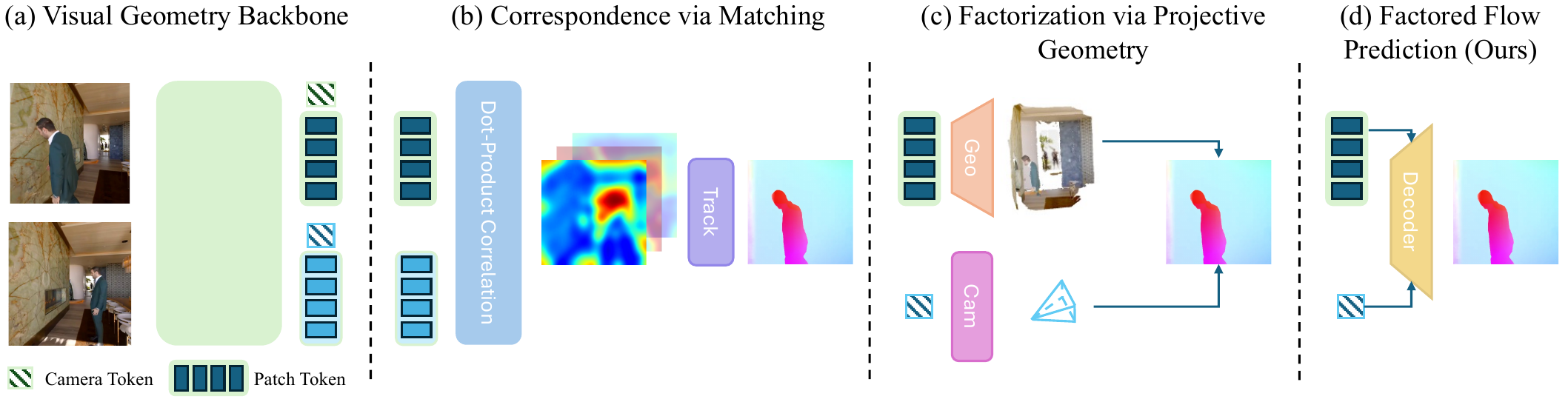}
    \vspace{-6mm}
    \caption{\textbf{Mechanisms for flow prediction.} (a) The visual geometry backbone first extracts the camera token and patch tokens of each input. (b) Existing correspondence heads~\cite{wang2025vggt} predict flow directly from local features via matching. (c) Flow can also be obtained by explicitly projecting predicted 3D points into another view via decoded camera parameters. However, this projection-based formulation is restricted to static scenes and sensitive to geometric errors. (d) Our factored flow approach conditions source-view geometry latents on the target-view camera latent and decodes dense correspondences directly in latent space, providing a geometry-aware and robust flow prediction mechanism that naturally extends to dynamic scenes.}
  \label{fig:flow_paradigm_comparison}
\end{figure*}

\vspace{1mm}
\noindent \textbf{Dense Correspondence Learning.}
Dense correspondence methods estimate pixel-level matches across images, providing a foundation for downstream camera motion and 3D structure recovery. These have progressed from classical optical flow~\cite{horn1981determining, lucas1981iterative} to recent wide-baseline matchers such as DKM~\cite{edstedt2023dkm}, RoMa~\cite{edstedt2024roma}, and UFM~\cite{zhang2025ufm} that achieve robust matching across large viewpoint changes. Video-based trackers like TAPIR and CoTracker~\cite{karaev2024cotracker,harley2022particle, doersch2022tap,doersch2023tapir} further extend correspondence estimation to long-range temporal reasoning across multiple frames. \ours{} builds on these models, using their predicted correspondences as training supervision to guide 3D geometry learning on data without 3D labels.

\vspace{1mm}
\noindent \textbf{Correspondence-driven Reconstruction.} Building on dense correspondence estimation, reconstruction methods recover 3D structure and camera motion from inferred matches. Classical Structure-from-Motion~\cite{schonberger2016structure} pipelines estimate camera poses and scene structure by detecting local features, computing pairwise correspondences, and jointly optimizing camera parameters and 3D points through bundle adjustment. Visual SLAM systems~\cite{mur2015orb,engel2017direct} extend this to dynamic settings, tracking features across frames to jointly estimate camera trajectories and scene geometry.
Recent approaches, including Robust-CVD~\cite{kopf2021robust}, CasualSAM~\cite{zhang2022structure}, VIPE~\cite{huang2025vipe} and MegaSAM~\cite{li2025megasam}, further incorporate monocular depth priors to enhance robustness under motion and occlusion.
However, these methods remain optimization-based, requiring per-video refinement and lacking feed-forward efficiency.

\vspace{1mm}
\noindent \textbf{Feed-forward Visual Geometry Learning.}
Recent efforts aim to replace traditional optimization pipelines with feed-forward networks that directly predict visual geometry from images. 
DUSt3R~\cite{wang2023DUSt3R} first demonstrated that dense pointmaps can be estimated from image pairs within a shared coordinate system, enabling efficient two-view reconstruction. 
MASt3R~\cite{leroy2024grounding} further improves this paradigm by introducing a learned matching head for better correspondence reasoning, while DiffusionSfM~\cite{zhao2025diffusionsfm} and VGGT~\cite{wang2025vggt} generalize to multi-view settings, jointly estimating camera parameters and scene structure. 
Subsequent works such as MonST3R~\cite{zhang2024monst3r}, CUT3R~\cite{wang2025continuous}, and StreamVGGT~\cite{zhuo2025streaming} extend this formulation to dynamic scenes, learning temporally consistent geometry across video frames. 
However, these models rely on 3D and camera supervision, which is not easily scalable. 
In contrast, our method enables scalable feed-forward learning of dynamic visual geometry using flow as auxiliary supervision, allowing training on unlabeled real-world videos.

\section{Approach}
\label{sec:train}

Our approach, \ours{}, scales visual geometry learning by supervising factored flow prediction, enabling the use of in-the-wild data without ground-truth geometry annotations (\eg, camera poses and depth maps). We first review the standard paradigm for fully supervised visual geometry learning (Sec.~\ref{sec:preliminaries}), then introduce our factored flow prediction formulation (Sec.~\ref{sec:factorization}), which decodes flow by combining the camera token from one view with the patch tokens from another -- allowing flow supervision to directly guide geometry learning while naturally extending to dynamic scenes. Finally, we describe the overall supervision signals and architecture (Sec.~\ref{sec:architecture}).

\vspace{-2pt}

\subsection{Preliminaries: Visual Geometry Networks}
\label{sec:preliminaries}

State-of-the-art visual geometry networks (\eg, VGGT~\cite{wang2025vggt} and $\pi^3$~\cite{wang2025pi}) take as input a set of images $\{I_1, I_2, \ldots, I_N\}$ and infer 3D scene geometry through a unified multi-view transformer. For instance, VGGT first encodes the input images into latent patch tokens using an off-the-shelf vision backbone~\cite{oquab2023dinov2}:
\begin{equation}
\mathbf{X}_i = f_{\text{enc}}(I_i), \quad \mathbf{X}_i \in \mathbb{R}^{P \times D},
\end{equation}
where $P$ and $D$ denote the number of patch tokens and their feature dimension, respectively.

A multi-view transformer then performs cross-view reasoning over these features to jointly infer camera-related features and geometry-related features:
\begin{equation}
\{\mathbf{c}_i,\, \mathbf{g}_i\}_{i=1}^{N} = f_{\text{multi-view}}\big(\{\mathbf{X}_i\}_{i=1}^{N}\big),
\label{eq:multi_view_transformer}
\end{equation}
where $\mathbf{c}_i$ and $\mathbf{g}_i$ denote the per-view camera features and geometry features, respectively
\footnote{In VGGT, $\mathbf{c}_i$ corresponds to a learnable camera token appended to patch tokens, whereas in $\pi^{3}$ it represents latent camera features predicted from the multi-view transformer rather than a single token.}.

Finally, the updated latent representations are decoded into explicit geometric properties -- camera poses, depths, and global pointmaps.

\begin{equation}
[\hat{\mathbf{R}}_i,\, \hat{\mathbf{t}}_i, \hat{\mathbf{K}}_i] = f_{\text{cam}}(\mathbf{c}_i),
\label{eq:camera_head}
\end{equation}
\begin{equation}
\hat{\mathbf{D}}_i,\, \hat{\mathbf{P}}_i = f_{\text{depth}}(\mathbf{g}_i), f_{\text{point}}(\mathbf{g}_i).
\label{eq:geometry_head}
\end{equation}
These outputs are supervised using ground-truth labels, often obtained from multi-view reconstruction pipelines such as Structure-from-Motion~\cite{agarwal2011building,schonberger2016structure}.
However, such supervision is expensive and limited in coverage, restricting the diversity of scenes and motions seen during training -- motivating our exploration of scalable alternatives.

\begin{figure*}[t]
  \centering
  \includegraphics[width=\linewidth]{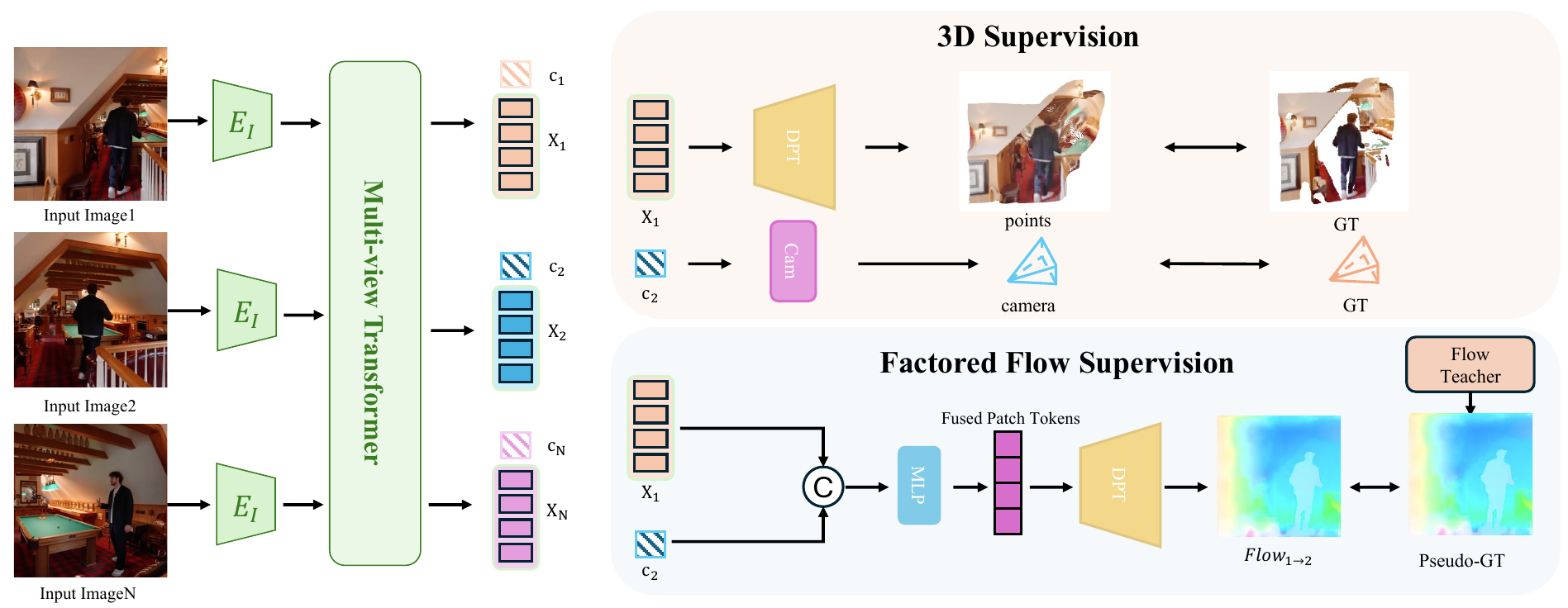}
  \vspace{-6mm}
  \caption{\textbf{Overview of \ours{}.} \ours{} predicts visual geometry using factored flow supervision, enabling scalable geometry learning from unlabeled videos. Each input image is encoded and processed by the multi-view transformer to produce camera tokens and patch tokens. For data with dense geometry and pose labels, we directly supervise the patch tokens and camera tokens with the corresponding labels. For unlabeled datasets without geometry and pose supervision, we predict flow between two frames in a factorized manner, supervised by the pseudo labels from an off-the-shelf 2D flow prediction model~\cite{zhang2025ufm}. To obtain the factored flow, we fuse the patch features of one frame with the camera features of another, and decode the fused representation through the DPT head to produce dense flow predictions.}
  \label{fig:method}
  \vspace{-2mm}
\end{figure*}

\subsection{Learning Visual Geometry via Factored Flow}
\label{sec:factorization}

\ours{} leverages \textit{flow} (dense pixel correspondences) as an additional supervision signal for visual geometry learning, enabling the use of in-the-wild data without ground-truth camera or depth annotations. The key technical question we answer is about \emph{how} one can effectively leverage such supervision, and we first detail different possible alternatives before describing our proposed mechanism:

\vspace{1mm}
\noindent \textbf{Flow as Visual Correspondence.} One possible approach is to infer flow using architecture relying on local feature `matching'  across image pairs (Fig.~\ref{fig:flow_paradigm_comparison} (b)). Indeed, prior multi-view models~\cite{wang2025vggt} adopt this design and leverage flow as an auxiliary  training signal. However, this design primarily ensures visually discriminative local features but, as we empirically show, does not facilitate the learning of scene geometry and camera motion.

\vspace{1mm}
\noindent \textbf{Flow from Explicit Camera and Scene Geometry.} 
Classical projective geometry provides an alternative mechanism to infer flow between images. Considering a static scene with predicted global pointmap $\hat{\mathbf{P}}_i$ from view $i$ and the camera parameters $(\hat{\mathbf R}_j, \hat{\mathbf t}_j, \hat{\mathbf K}_j)$ for view~$j$, the flow between these images can be analytically computed by projecting $\hat{\mathbf{P}}_i(\mathbf{u}_i)$ 
into camera~$j$:
\begin{equation}
\hat{\mathbf{F}}_{i\!\rightarrow\!j}(\mathbf{u}_i) = \hat{\mathbf{u}}_{i\!\rightarrow\!j} = 
\pi\!\big(\hat{\mathbf{K}}_j (\hat{\mathbf{R}}_j\,\hat{\mathbf{P}}_i(\mathbf{u}_i) + \hat{\mathbf{t}}_j)\big),
\label{eq:projection}
\end{equation}
where the operator $\pi(\cdot)$ denotes perspective projection from 3D points to pixel coordinates.

This explicit flow computation mechanism, grounded in projective geometry, provides an ideal factorization that jointly promotes learning of both scene geometry and camera motion, as illustrated in Fig.~\ref{fig:flow_paradigm_comparison} (c).
However, the pure projection in Eq.~\ref{eq:projection} cannot account for scene motion. While this can be mitigated by addition scene flow prediction as in ZeroMSF~\cite{liang2025zeroshot}, we find that supervising flow from the decoded cameras and depths can introduce instability even in static scenes, as errors in either component may lead to large inaccuracies in the resulting flow.

\vspace{1mm}
\noindent \textbf{Factored Flow Prediction.} 
Our approach builds on an insight from the projection-based flow computation -- that asymmetric flow computation using geometry from the source and pose from the target can better guide visual geometry learning. However, rather than relying on analytical computation, we propose to leverage a learned \emph{factored flow prediction} mechanism to infer flow from pose and geometry latent representations (see (d) in Fig.~\ref{fig:flow_paradigm_comparison}).

More specifically, given the geometry latents $\mathbf{g}_i$ from view~$i$ and camera latents $\mathbf{c}_j$ from view~$j$ (the outputs of the multi-view transformer in Eq.~\ref{eq:multi_view_transformer}), \ours{} leverages a flow prediction head to predict the flow:
\begin{equation} \hat{\mathbf{F}}_{i\!\rightarrow\!j} = \Phi_{\text{flow}}(\mathbf{g}_i, \mathbf{c}_j),
\label{eq:flow_predicion}
\end{equation}
where $\Phi_{\text{flow}}$ denotes a learned flow prediction module. The camera latents from view~$j$ are used to modulate the geometry latents from view~$i$, which are then decoded by a DPT head~\cite{ranftl2021vision} to produce the flow field.

This approach bypasses the need to decode explicit geometric elements for flow computation, improving robustness and enabling end-to-end training. Moreover, it implicitly handles dynamic scenes where the flow field reflects a combination of camera motion and scene motion rather than pure geometric projection.

\begin{table*}[t!]
\centering
\small
\resizebox{\linewidth}{!}{
\begin{tabular}{llcccc@{\hspace{18pt}}lcccc}
\toprule
& \multicolumn{5}{c}{\textbf{Static Scenes}} & \multicolumn{5}{c}{\textbf{Dynamic Scenes}}\\
\cmidrule(lr){2-6}\cmidrule(lr){7-11}
\textbf{Model Variant} & \textbf{Data (\# Seqs)} & \textbf{RRA@30}$\uparrow$ & \textbf{RTA@30}$\uparrow$ & \textbf{CD}$\downarrow$ & \textbf{MSE}$\downarrow$
&
\textbf{Data (\# Seqs)} & \textbf{RRA@30}$\uparrow$ & \textbf{RTA@30}$\uparrow$ & \textbf{CD}$\downarrow$ & \textbf{MSE}$\downarrow$\\
\midrule
\expa{} & ScanNet++ ($1$K) & 0.7500 & 0.6929 & 0.030 & 0.088
&
OmniWorld ($1$K) & 66.01 & 62.37 & 0.105 & 0.637\\
\midrule
\expb{} & \textbf{+} ScanNet++ ($1$K) & 0.6700 & 0.4572 & 0.033 & 0.088
&
\textbf{+} SpatialVID ($3$K) & 61.23 & 56.12 & 0.158 & 0.710\\
\expc{} & \textbf{+} ScanNet++ ($1$K) & 0.7438 & 0.7021 & 0.030 & 0.089
&
\textbf{+} SpatialVID ($3$K) & 68.56 & 62.95 & 0.107 & 0.628\\
\expd{} & \textbf{+} ScanNet++ ($1$K) & \textbf{0.7700} & \textbf{0.7366} & \textbf{0.026} & \textbf{0.078}
&
\textbf{+} SpatialVID ($3$K) & \textbf{76.26} & \textbf{68.84} & \textbf{0.103} & \textbf{0.598}\\
\bottomrule
\end{tabular}
}
\caption{\textbf{Does factored flow prediction help visual geometry learning?}
We evaluate the camera and geometry prediction of a model trained with only 3D labeled data and variants that use additional unlabeled data via flow supervision. We find that on both static and dynamic scenes, our proposed factored flow prediction (\expd{}) yields consistent gains over the model variant without additional unlabeled data (\expa{}) as well as other flow-supervised alternatives.
}
\label{tab:static_dynamic_merged}
\vspace{-8pt}
\end{table*}

\begin{figure*}[t!]
  \centering
  \includegraphics[width=\linewidth]{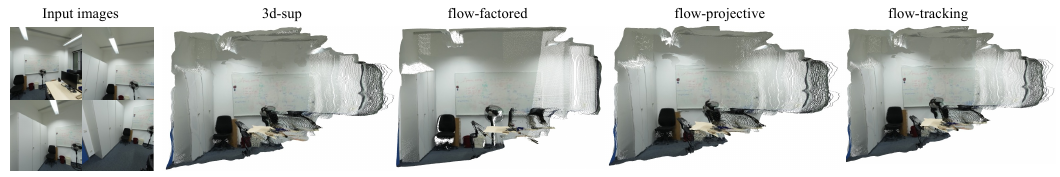}
  \vspace{-6mm}
  \caption{\textbf{Factored flow prediction aids visual geometry learning.} Compared with the baseline (\expa{}) and alternative formulations that use flow supervision (\expb{}, \expc{}), \ours{} (\expd{}) yields more accurate scene reconstruction, highlighting the benefits of unlabeled data and our factored flow formulation.} 
  \label{fig:flow_formulation_comparison}
\end{figure*}

\begin{figure*}[t!]
  \centering
  \includegraphics[width=\linewidth]{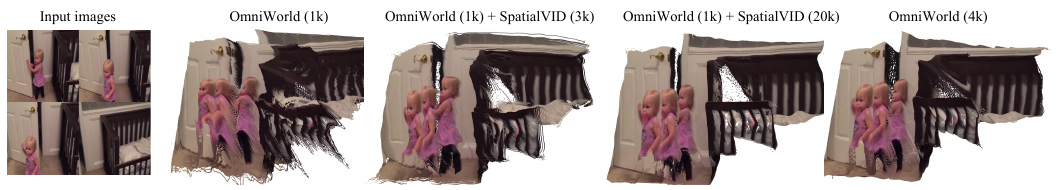}
  \vspace{-6mm}
  \caption{\textbf{Scaling with unlabeled videos.} With factored flow supervision, increasing the amount of unlabeled SpatialVID data 
progressively improves dynamic-scene reconstruction quality, even surpassing the model with additional 3D-labeled training data.} 
  \label{fig:flow_scaling_comparison}
\end{figure*}

\subsection{Overall Architecture and Learning Objectives}
\label{sec:architecture}
\begingroup
\ours{} builds upon a standard supervised visual geometry learning framework (Section \ref{sec:preliminaries}) -- a multi-view transformer backbone with camera and geometry heads supervised using labeled 3D data. In addition, \ours{} learns a factored flow prediction module, allowing 2D flow as supervision on both labeled and (crucially) unlabeled data, enabling scalable visual geometry learning across both static and dynamic scenes -- see  Fig.~\ref{fig:method} for an overview.

\vspace{3mm}
\noindent \textbf{Camera and Geometry Supervision from Labeled Data.} 
\ours~ predicts per-image cameras and local geometry along with confidence maps that indicate model uncertainty. 
When 3D labels are available, we supervise the decoded predictions using ground-truth camera poses $\mathbf{T}_i$ and geometry annotations $\mathbf{G}_i$:
\begin{equation}
\mathcal{L}_{sup} =
\sum_{i=1}^{N}
\left(
\lambda_{cam}\,\mathcal{L}_{cam}(\hat{\mathbf{T}}_i, \mathbf{T}_i)
+
\lambda_{geo}\,\mathcal{L}_{geo}(\hat{\mathbf G}_i, \mathbf G_i)
\right).
\end{equation}

To disentangle visual geometry learning from predefined coordinate frames, \ours{} adopts the permutation-equivariant geometry prediction framework introduced by $\pi^3$~\cite{wang2025pi}, which predicts geometry in local coordinates without relying on a fixed global frame. Models that predict geometry in a global frame (\eg, VGGT~\cite{wang2025vggt}) can be modified to follow the same design by removing the `first camera' token and using an optimal-alignment-based loss for supervising the predicted pointmaps $\hat{\mathbf P}$

(see appendix for details):
\begin{equation}
\mathcal{L}_{\text{point}}
=
\frac{1}{K}
\min_{\mathbf{R}\in SO(3),\,\mathbf{t}\in\mathbb{R}^3}
\sum_{k}
\left\|
\mathbf{R}\hat{\mathbf p}_k + \mathbf{t} - \mathbf p_k
\right\|_2 ,
\end{equation}
where $\hat{\mathbf p}_k \in \hat{\mathbf P}$ and $\mathbf p_k \in \mathbf P$ denote corresponding 3D points in a sequence, and $K$ is the number of valid points.

\vspace{3mm}
\noindent \textbf{Flow Supervision from Labeled and Unlabeled Data.} We supervise the proposed flow head with a robust regression loss~\cite{edstedt2024roma} between the predicted coordinates $\hat{\mathbf{u}}_{i\!\rightarrow\!j}$ and the ground-truth correspondences $\mathbf{u}_{i\!\rightarrow\!j}$:
\begin{equation*}
\mathcal{L}_{\text{flow}} = 
\frac{1}{\sum_{p \in \Omega} \mathbf{C}[p]} 
\sum_{p \in \Omega} \mathbf{C}[p] \,
\ell_{\text{robust}}\!\left(
\|\hat{\mathbf{u}}_{i\!\rightarrow\!j}[p] - \mathbf{u}_{i\!\rightarrow\!j}[p]\|_2
\right),
\end{equation*}
where $p \in \Omega$ indexes pixels, $\mathbf{C}$ denotes the covisibility mask between view~$i$ and $j$, and $\ell_{\text{robust}}$ is a generalized Charbonnier loss emphasizing inliers. We use this loss on both labeled and unlabeled data,  relying on pseudo-ground-truth 2D correspondences from a pre-trained teacher (UFM~\cite{zhang2025ufm}) when ground-truth 2D flow is unavailable.

\section{Experiments}
We evaluate \ours{} in two stages. First, in controlled experiments, we show that factored flow prediction outperforms alternative flow-based designs and that performance scales consistently with the amount of unlabeled data (Sec.~\ref{sec:abla1}). We then integrate factored flow prediction into state-of-the-art visual geometry networks and demonstrate improvements across eight benchmarks by leveraging ${\sim}800$K unlabeled videos (Sec.~\ref{sec:fullmodel}).

\begin{figure*}[t]
  \centering
  \includegraphics[width=\linewidth]{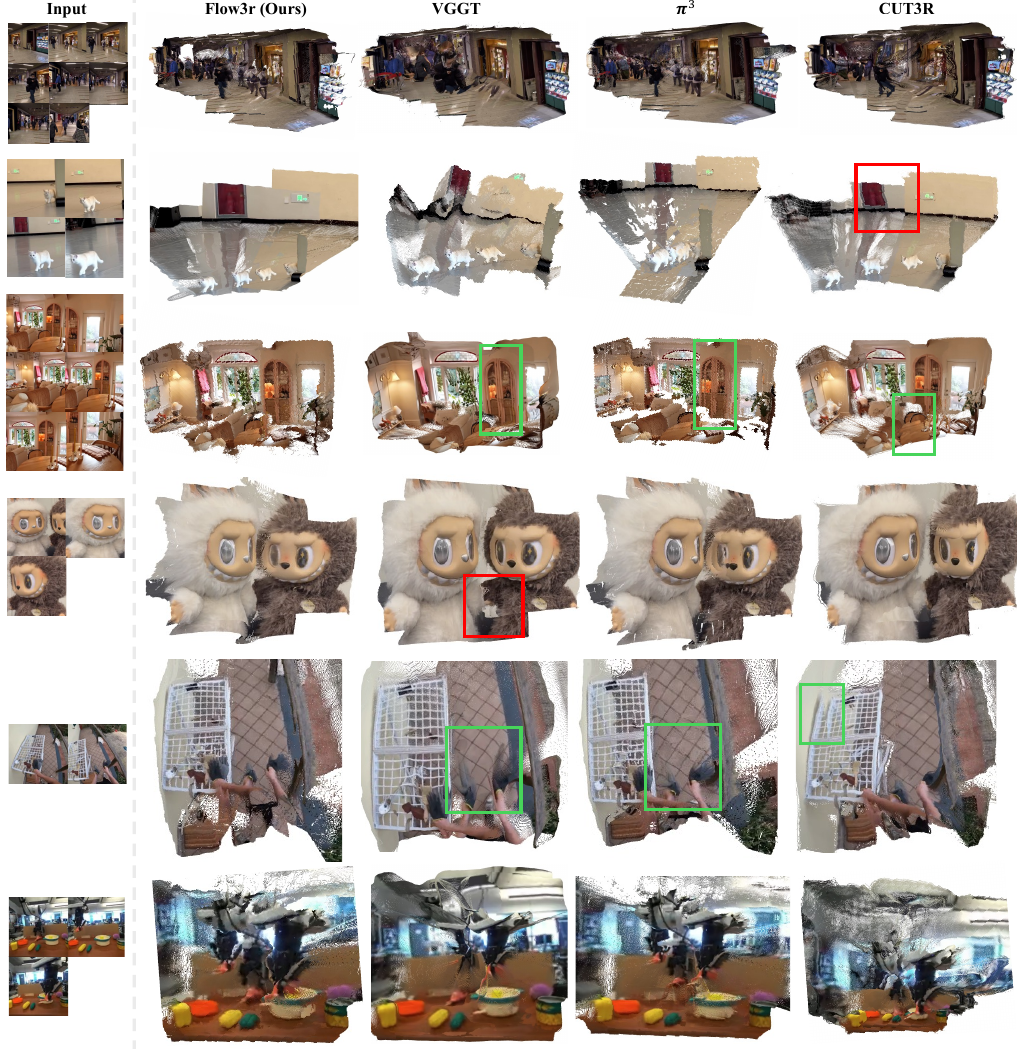}
  \vspace{-6mm}
  \caption{\textbf{Qualitative results on in-the-wild data.} We compare the reconstructions from \ours{} with representative feedforward reconstruction methods on in-the-wild multi-view images from static scenes, dynamic scenes, and interaction videos. We observe that \ours{} infers cleaner and more accurate scene structure than baselines that often yield misaligned outputs or infer inaccurate motion.}
  \label{fig:main_comparison}
\end{figure*}

\setlength{\tabcolsep}{4pt} 
\begin{table*}[t]
\centering
\footnotesize
\resizebox{\textwidth}{!}{%
\begin{tabular}{l cccccccccccccccc}
\toprule
\multirow{2}{*}{\textbf{Methods}} &
\multicolumn{4}{c}{\textbf{Kinetics700}} &
\multicolumn{4}{c}{\textbf{EPIC-KITCHENS}} &
\multicolumn{4}{c}{\textbf{Sintel}} &
\multicolumn{4}{c}{\textbf{Bonn}} \\
\cmidrule(lr){2-5} \cmidrule(lr){6-9} \cmidrule(lr){10-13} \cmidrule(lr){14-17}
& \textbf{RPE trans}$\downarrow$ & \textbf{RPE rot}$\downarrow$ & \textbf{MSE}$\downarrow$ & \textbf{f-score@5}$\uparrow$ & 
  \textbf{RPE trans}$\downarrow$ & \textbf{RPE rot}$\downarrow$ & \textbf{MSE}$\downarrow$ & \textbf{f-score@5}$\uparrow$ & 
  \textbf{RPE trans}$\downarrow$ & \textbf{RPE rot}$\downarrow$ & \textbf{MSE}$\downarrow$ & \textbf{f-score@5}$\uparrow$& 
  \textbf{RPE trans}$\downarrow$ & \textbf{RPE rot}$\downarrow$ & \textbf{MSE}$\downarrow$ & \textbf{f-score@5}$\uparrow$   \\
\midrule
DUSt3R &0.063&9.343& 0.366 & 0.533&
0.110 & 8.492 & 0.312 & 0.528&
0.179&15.166 & 0.622 & 0.271&
0.113 & 6.384& 0.116 & 0.800 \\

CUT3R &\third{0.027} &1.988&\third{0.303} & \third{0.573}&
0.081 & 4.709 & 0.338 & 0.493&
0.128  &1.998& 0.676 & 0.217&
\second{0.095} & 6.349 & 0.088 & \third{0.899} \\

VGGT &0.038&\third{1.392}&0.347&0.479&
\third{0.049}&\second{3.025}&\third{0.220}&\third{0.617}&
\third{0.086}&\third{1.220}&\third{0.595}&\third{0.311}&
0.096&\best{6.021}&\third{0.082}&0.884\\

$\pi^{3}$ & \second{0.023}&\second{1.006}&\second{0.267}&\second{0.585}&
\second{0.043}&\second{3.025}&\second{0.200}&\second{0.620}&
\second{0.066}&\second{1.122}&\second{0.523}&\third{0.317}& \second{0.095}&\second{6.240}&\second{0.076}&\second{0.905}\\

Flow3r
&\best{0.018}&\best{0.830}&\best{0.256}&\best{0.599}&
\best{0.037}&\best{2.729}&\best{0.199}&\best{0.622}&
\best{0.058}&\best{0.920}&\best{0.426}&\best{0.404}&
\best{0.094}&\third{6.262}&\best{0.052}&\best{0.954} \\

\bottomrule
\end{tabular}
}
\vspace{-5pt}
\caption{\textbf{Comparison on dynamic datasets.} We evaluate the camera (RPE rot and trans) and geometry prediction (MSE and f-score) for dynamic scene reconstruction across different datasets.
Best, second and third results are highlighted in light red, orange, and yellow, respectively. \ours{} consistently outperforms state-of-the-art methods in both camera pose estimation and scene reconstruction, demonstrating the effectiveness of leveraging large-scale unlabeled videos for visual geometry learning via factored-flow supervision.}
\label{tab:result_dynamic}
\end{table*}

\setlength{\tabcolsep}{4pt} 
\begin{table*}[t]
\centering
\footnotesize
\resizebox{\textwidth}{!}{%
\begin{tabular}{l cccccccccccccccc}
\toprule
\multirow{2}{*}{\textbf{Methods}} &
\multicolumn{4}{c}{\textbf{Co3Dv2}} &
\multicolumn{4}{c}{\textbf{Scannet}} &
\multicolumn{4}{c}{\textbf{NRGBD}} &
\multicolumn{4}{c}{\textbf{7-scenes}} \\
\cmidrule(lr){2-5} \cmidrule(lr){6-9} \cmidrule(lr){10-13} \cmidrule(lr){14-17}
& \textbf{RTA@30}$\uparrow$ & \textbf{AUC@30}$\uparrow$ & \textbf{MSE}$\downarrow$ & \textbf{f-score@5}$\uparrow$ & 
  \textbf{RTA@30}$\uparrow$ & \textbf{AUC@30}$\uparrow$& \textbf{MSE}$\downarrow$ & \textbf{f-score@5}$\uparrow$ & 
  \textbf{RTA@30}$\uparrow$ & \textbf{AUC@30}$\uparrow$& \textbf{MSE}$\downarrow$ & \textbf{f-score@5}$\uparrow$& 
  \textbf{RTA@30}$\uparrow$ & \textbf{AUC@30}$\uparrow$& \textbf{MSE}$\downarrow$ & \textbf{f-score@5}$\uparrow$   \\
\midrule
DUSt3R &90.68 &68.11& 0.223 & 0.783&
57.14 &31.56 & 0.092 & 0.831&
93.25&76.04 & 0.063 & 0.865&
76.59 & 55.07& \third{0.170} & \third{0.714} \\

CUT3R &90.68 &68.11& 0.278 &0.737& 71.39& 42.30 & 0.132 &0.740 &95.63 &76.90& 0.075&0.808&81.15& 59.72&0.183&0.695 \\

VGGT &\third{97.27}&\third{87.62}&\second{0.151}&\second{0.874}&
\best{93.71}&\best{71.37}&\best{0.053}&\second{0.931}&\second{99.21}&\second{93.13}&\third{0.032}&\third{0.964}&
\third{86.51}&\third{69.77}&0.182&0.665\\

$\pi^{3}$ & \second{97.49}&\best{90.53}&\second{0.151}&\second{0.874}&
\third{91.14}&\third{69.39}&\third{0.058}&\third{0.930}&
\third{99.20}&\third{90.40}&\second{0.021}&\second{0.983}& \second{87.69}&\second{71.80}&\second{0.169}&\second{0.737}\\

Flow3r
&\best{97.62}&\second{90.41}&\best{0.150}&\best{0.876}&
\second{92.89}&\second{71.00}&\best{0.053}&\best{0.943}&
\best{99.60}&\best{94.40}&\best{0.018}&\best{0.992}&
\best{91.66}&\best{75.76}&\best{0.102}&\best{0.807} \\
\bottomrule
\end{tabular}
}
\vspace{-5pt}
\caption{\textbf{Comparison on static datasets.} We evaluate the camera (RTA and AUC) and geometry prediction (MSE and f-score) for static scene reconstruction from the multi-view input. Best, second and third results are highlighted in light red, orange, and yellow, respectively. \ours{} outperforms state-of-the-art methods, indicating that the gains from incorporating unlabeled dynamic videos for visual geometry learning also transfer to static scenes.}
\label{tab:result_static}
\vspace{-10pt}
\end{table*}

\begin{table}[h]
\centering
\small
\resizebox{\linewidth}{!}{
\begin{tabular}{llcccc}
\toprule
\multicolumn{2}{c}{\textbf{Data}} & \multicolumn{4}{c}{\textbf{Metrics}} \\
\cmidrule(lr){1-2}\cmidrule(lr){3-6}
\textbf{Labeled (\# Seqs)} & \textbf{Unlabeled (\# Seqs)} & \textbf{RRA@30}$\uparrow$ & \textbf{RTA@30}$\uparrow$ & \textbf{CD}$\downarrow$ & \textbf{MSE}$\downarrow$ \\
\midrule
OmniWorld ($1$K) & \multicolumn{1}{c}{-} & 66.01 & 62.37 &0.105&0.637\\
\midrule
OmniWorld ($1$K) &  SpatialVID ($3$K) & 76.26 & 68.84 & 0.103 & 0.598\\
OmniWorld ($1$K) &   SpatialVID ($10$K) & 78.45 & 68.82&\underline{0.077}& \underline{0.560}\\
OmniWorld ($1$K) &   SpatialVID ($20$K) & \textbf{81.12} & \textbf{71.21} & \textbf{0.075} & \textbf{0.532} \\
\midrule
OmniWorld ($4$K)  &  \multicolumn{1}{c}{-} & \underline{78.68} & \underline{70.26} & 0.080&0.565 \\
\bottomrule
\end{tabular}}
\vspace{-8pt}
\caption{\textbf{Scaling with unlabeled videos for dynamic visual geometry learning.}
We train model variants on OmniWorld~\cite{zhou2025OmniWorld} (labeled 3D supervision) together with increasing amounts of SpatialVID~\cite{wang2025spatialvid} (unlabeled videos with flow supervision).
Performance improves consistently as we scale up the unlabeled data.}
\vspace{-5pt}
\label{tab:ablation_scaling}
\end{table}

\begin{table}[ht]
    \centering
    \small
    \setlength{\tabcolsep}{4pt}
    \resizebox{\linewidth}{!}{
    \begin{tabular}{l c c c c c c c c c c c}
        \toprule
        \multirow{2}{*}{\textbf{Base}} &
        \multirow{2}{*}{\textbf{3D Sup.}} &
        \multicolumn{2}{c}{\textbf{Flow Loss}} &
        \multicolumn{4}{c}{\textbf{EPIC-KITCHENS}} &
        \multicolumn{4}{c}{\textbf{7-scenes}} \\
        \cmidrule(lr){3-4} \cmidrule(lr){5-8} \cmidrule(lr){9-12}
        & & \textbf{Labeled} & \textbf{Unlabeled} &
        \textbf{trans}$\downarrow$ & \textbf{rot}$\downarrow$ & \textbf{MSE}$\downarrow$ & \textbf{f-score}$\uparrow$ &
        \textbf{RTA}$\uparrow$ & \textbf{AUC}$\uparrow$ & \textbf{MSE}$\downarrow$ & \textbf{f-score}$\uparrow$ \\
        \midrule

        \multirow{4}{*}{VGGT} &
        \multicolumn{3}{c}{N/A (base model)} &
        0.049 & 3.025 & 0.220 & 0.617 &
        86.51&69.77&0.182&0.665 \\
        & \ding{51} & \ding{55} & \ding{55} &
        0.050 & 3.005 & 0.218 & 0.530 &
        88.10 & 68.23&0.167&\textbf{0.750}
         \\
        & \ding{51} & \ding{51} & \ding{55} &
        0.045 & 3.005 & 0.203 & 0.530 &
        89.17&68.22&0.167&0.749
         \\
        & \ding{51} & \ding{51} & \ding{51} &
        \textbf{0.039} & \textbf{2.808} & \textbf{0.202} & \textbf{0.599} &
        \textbf{90.67} & \textbf{71.75} & \textbf{0.166} & \textbf{0.750} \\
        \midrule

        \multirow{4}{*}{$\pi^3$} &
        \multicolumn{3}{c}{N/A (base model)} &
        0.043 & 3.025 & 0.200 & 0.620 &
        87.69&71.80&0.169&0.737 \\
        & \ding{51} & \ding{55} & \ding{55} &
        0.041 & 3.005 & 0.218 & 0.620 &
        87.27&72.66&0.169&0.735
         \\
        & \ding{51} & \ding{51} & \ding{55} &
        0.040 & 3.005 & 0.203 & 0.619 &
        87.99&72.47&0.149&0.783
         \\
        & \ding{51} & \ding{51} & \ding{51} &
        \textbf{0.037} & \textbf{2.729} & \textbf{0.199} & \textbf{0.622} & \textbf{91.66}&\textbf{75.76}&\textbf{0.102}&\textbf{0.807} \\

        \bottomrule
    \end{tabular}}
    \vspace{-8pt}
    \caption{\textbf{Effect of adding unlabeled videos via flow supervision.}
    For each backbone (VGGT~\cite{wang2025vggt} and $\pi^3$~\cite{wang2025pi}), we compare a 3D supervision only baseline, a labeled multi-task model (geometry + flow), and the full model incorporating additional unlabeled data via flow supervision. This demonstrates the effectiveness of unlabeled data via flow supervision for visual geometry learning.}
\label{tab:result_dynamic_3}
\vspace{-10pt}
\end{table}

\subsection{Flow Supervision Improves Visual Geometry}
\label{sec:abla1}
We first evaluate whether factored flow prediction improves visual geometry learning in a controlled setting. We compare against training with only labeled 3D data, and against alternative flow prediction designs.
We also analyze the scaling behavior by increasing the unlabeled data.

\vspace{1mm}
\noindent \textbf{Experimental Designs.} We denote the `base model' trained with only 3D supervision as \expa{}. Building upon this `no-flow' baseline, we study three variants that incorporate \emph{additional} unlabeled data via flow supervision using different formulations: (1) \expb{},  computes flow explicitly from predicted camera poses and pointmaps via projective geometry; (2) \expc{},  adopts a VGGT-style~\cite{wang2025vggt} tracking head based on pairwise patch features; (3) \expd{}, applies our proposed factored flow prediction formulation. All models share a (small) VGGT-like architecture, and are trained \emph{from scratch} (except for the image encoder initialized from DINO~\cite{oquab2023dinov2}) -- see the appendix for details.

\noindent \emph{Static Scenes.} We train and evaluate on ScanNet++~\cite{yeshwanth2023ScanNet++} using $1K$ scenes as sources of labeled 3D data and another $1K$ scenes as `unlabeled'. The model variants with flow supervision apply flow loss only to the additional unlabeled training sequences, ensuring that all model variants (except \expa{}) are trained with the same total number of sequences. 

\noindent \emph{Dynamic Scenes.} We use the OmniWorld~\cite{zhou2025OmniWorld} dataset for 3D supervision and evaluation. As a scalable source of unlabeled dynamic data, we incorporate videos from SpatialVID~\cite{wang2025spatialvid} for flow supervision, using UFM~\cite{zhang2025ufm} to extract the pseudo-ground-truth 2D flow labels. 

\noindent We evaluate both camera pose and geometry predictions. Relative Rotation Accuracy (RRA) and Relative Translation Accuracy (RTA) measure the accuracy of predicted camera rotations and translations, while Chamfer Distance (CD) and Mean-Square Error (MSE) assess the geometric correctness of the reconstructed 3D structure.

\vspace{1mm}
\noindent \textbf{Comparisons of Flow Prediction  Mechanisms.} The results in Tab.~\ref{tab:static_dynamic_merged} show that, compared to the base model \expa, our \expd{} model yields significant improvements in camera and geometry prediction.
We find that \expd{} also outperforms both flow-prediction alternatives (\expb{} and \expc{}), achieving higher camera pose accuracy and geometric quality in both static and dynamic scenes. Notably, even though \expc{} learns to predict accurate flow (see the appendix), it provides almost no improvement in pose accuracy and geometric quality, suggesting that supervising flow prediction from pairwise patch features may help the network to learn discriminative features, but does not meaningfully benefit visual geometry learning. Moreover, \expb{} even degrades performance, indicating that supervising flow computed from explicit camera and geometry predictions may suffer from instability and thus harm learning. Qualitatively, Fig.~\ref{fig:flow_formulation_comparison} shows that \texttt{flow-factored} yields cleaner reconstructions than \texttt{3d-sup} while also improving over other flow mechanisms. Although our factored flow prediction is \emph{suboptimal} for standalone flow estimation -- since it enforces an information bottleneck by conditioning on the target-view camera token rather than patch features that contain richer visual cues (see appendix for evaluation and analysis) -- these results demonstrate that it provides a more effective supervisory signal for visual geometry learning.

\vspace{1mm}
\noindent \textbf{Scaling with Unlabeled Data.}
To investigate the scaling behavior of the proposed factored flow prediction, we progressively increase the number of unlabeled dynamic videos used for flow supervision on SpatialVID. Specifically, we keep the 3D-labeled OmniWorld set fixed (1K sequences) and scale the number of SpatialVID sequences used to apply the flow loss (3K, 10K, and 20K sequences). For reference, we also train a model using only additional labeled data by increasing OmniWorld to 4K sequences without any unlabeled videos. The results in Tab.~\ref{tab:ablation_scaling} show that scaling the total number of training sequences to a larger regime (\eg, 10$\times$ or 20$\times$ the amount used in the \expa{} no-flow baseline) yields consistent improvements. Notably, using 20K unlabeled videos together with 1K labeled sequences outperforms training with 4K labeled sequences alone. Qualitatively, Fig.~\ref{fig:flow_scaling_comparison} exhibits the same trend: increasing the amount of unlabeled flow supervision leads to visibly cleaner and more complete reconstructions, reinforcing the quantitative gains. These results confirm that factored flow prediction enables visual geometry learning to scale effectively with large amounts of unlabeled video data.

\subsection{Improving SoTA Models with Unlabeled Data}
\label{sec:fullmodel}

We demonstrated in Sec.~\ref{sec:abla1} that our factored flow supervision consistently improves visual geometry learning, motivating us to explore its efficacy in improving state-of-the-art models (\eg, $\pi^{3}$~\cite{wang2025pi}) by incorporating a flow prediction module and leveraging large-scale unlabeled dynamic video data as auxiliary supervision.

\vspace{1mm}
\noindent \textbf{Training Overview.}
Our model training proceeds in two stages. First, we initialize our model from pretrained visual geometry model checkpoint and append our factored flow prediction head to the model. In this stage, we keep the backbone frozen and use all labeled data to train only the newly added flow head (\ie, the gradients from flow supervision do not update the visual geometry backbone). In the second stage, we add additional unlabeled video datasets for training, and unfreeze the whole model, performing end-to-end finetuning using both labeled and unlabeled data.

\vspace{1mm}
\noindent \textbf{Datasets.} We finetune the base model on a diverse mixture of labeled 3D/4D datasets and large-scale unlabeled video data.
Our labeled supervision set consists of eleven widely-used multi-view reconstruction datasets -- CO3Dv2~\cite{reizenstein2021common}, Habitat~\cite{savva2019habitat}, ARKitScenes~\cite{baruch2021arkitscenes}, ScanNet~\cite{dai2017scannet}, ScanNet++~\cite{yeshwanth2023ScanNet++}, MegaDepth~\cite{li2018megadepth}, BlendedMVS~\cite{li2018megadepth}, StaticThings3D~\cite{schroppel2022benchmark},
Omniworld~\cite{zhou2025OmniWorld},
PointOdyssey~\cite{zheng2023point}, and
VKITTI~\cite{gaidon2016virtual} -- which provide ground-truth camera poses and geometry for supervised learning, including approximately 34K sequences in total.
To further scale correspondence supervision, we incorporate three unlabeled 4D video datasets: Kinetics-700~\cite{carreira2019short}, SpatialVID~\cite{wang2025spatialvid} (high-quality dynamic sequences), and EPIC-Kitchens~\cite{damen2018scaling}. These datasets together contribute approximately 800K video sequences, offering substantial appearance and motion diversity for learning from unlabeled videos.

\vspace{1mm}
\noindent \textbf{Baselines and Metrics.} 
Our primary model, denoted as \ours, is built upon the $\pi^3$~\cite{wang2025pi} backbone. 
We compare \ours{} with representative feed-forward visual geometry baselines: CUT3R~\cite{wang2025continuous}, VGGT~\cite{wang2025vggt}, and $\pi^3$~\cite{wang2025pi}. We evaluate performance using pose accuracy and reconstruction metrics on four dynamic datasets using sparsely-sampled video frames as input: Kinetics700~\cite{carreira2019short}, Epic-Kitchens~\cite{damen2018scaling}, Sintel~\cite{butler2012naturalistic}, and Bonn~\cite{palazzolo2019refusion}. For Kinetics700 and Epic-Kitchens, we compute pseudo ground truth from dense videos using MegaSAM~\cite{li2025megasam}. We also evaluate on four static datasets: CO3Dv2~\cite{reizenstein2021common}, ScanNet~\cite{dai2017scannet}, NRGBD~\cite{azinovic2022neural}, and 7-Scenes~\cite{shotton2013scene}. Following prior work~\cite{chen2024leap, zhang2024monst3r}, we report Relative Pose Error, including RPE (trans) and RPE (rot) for dynamic datasets, and RTA and AUC for static datasets. We assess 3D geometry using mean squared error (MSE) between the predicted and ground-truth points, which captures overall geometric fidelity, and F-score to evaluate the accuracy-completeness trade-off. 

\vspace{1mm}
\noindent \textbf{Results.} 
We report quantitative results in Tab.~\ref{tab:result_dynamic} and Tab.~\ref{tab:result_static}, comparing \ours{} against state-of-the-art feed-forward visual geometry prediction models. For both
pose estimation and scene reconstruction, \ours{} consistently outperforms baselines. We see particularly large improvements for in-the-wild dynamic videos, where our unlabeled data provides additional supervisory signals, but also note an improvement on static datasets, indicating better generalization via large-scale supervision.

We also include qualitative results on in-the-wild multi-view images of static scenes, interaction videos and dynamic scenes in Fig.~\ref{fig:main_comparison}. We observe that \ours{} infers cleaner and more accurate scene structure than baselines, which often produce misaligned outputs (\eg, duplicated static components in room reconstruction) and fail to capture the motion of dynamic components (\eg, incorrect movement of the cat).

\vspace{1mm}
\noindent \textbf{Ablating the Effects of Unlabeled Data.}
Our main results show clear improvements on the pre-trained model using our framework, but these conflate additional finetuning and multi-task training on labeled data with the incorporation of unlabeled data. We perform additional experiments to disentangle these effects.
First, we finetune the base model using only labeled data without the flow head, establishing a camera-and-geometry-only baseline.
Second, we attach the flow head and train with labeled supervision only, forming a multi-task training setup that jointly optimizes camera, geometry and flow.
Comparing these two settings measures the impact of multi-task learning itself, while comparing the labeled-only multi-task model with the full \ours{} reveals the benefit of incorporating unlabeled videos.
To verify that this observation is not architecture-specific, we additionally repeat the same training procedure using the VGGT~\cite{wang2025vggt} backbone.
All settings use the same total number of training iterations for a fair comparison.
Tab.~\ref{tab:result_dynamic_3} shows that while additional finetuning and multi-task training on labeled data improves performance, the primary performance gains stem from incorporating unlabeled videos via flow supervision.
\section{Discussion}

In this work, we present \ours{} and demonstrate that it effectively leverages in-the-wild unlabeled data by introducing factored flow prediction, advancing visual geometry learning beyond existing fully supervised methods. While our approach opens up new possibilities, several challenges remain.
First, \ours{} relies on off-the-shelf models to provide pseudo-ground-truth flow supervision, and there can be domains where such 2D prediction fails. Second, although our factored flow formulation elegantly handles dynamic scenes and enables flow supervision to improve the learning of both camera motion and scene geometry, \ours{} may struggle under complex scenes with multiple independently moving components. 
Finally, our current experiments operate at a moderate scale ($\sim\!$800K video sequences for flow supervision), and scaling to truly large-scale settings ($\sim$10-100M videos) presents an exciting but unexplored direction, and we envision \ours's formulation serving as a building block for future large-scale learning methods.

\vspace{2mm}
\noindent \textbf{Acknowledgements.} We thank the members of the Physical Perception Lab at CMU for their valuable discussions. 
This work was supported by an NVIDIA academic grant. This work used Bridges-2~\cite{brown2021bridges} at Pittsburgh Supercomputing Center through allocation CIS250061 from the Advanced Cyberinfrastructure Coordination Ecosystem: Services \& Support (ACCESS) program, which is supported by
National Science Foundation grants \#2138259, \#2138286,
\#2138307, \#2137603, and \#2138296. This work was supported by Intelligence Advanced Research Projects Activity (IARPA) via Department of Interior/Interior Business
Center (DOI/IBC) contract number 140D0423C0074. The
U.S. Government is authorized to reproduce and distribute
reprints for Governmental purposes notwithstanding any
copyright annotation thereon. Disclaimer: The views and
conclusions contained herein are those of the authors and
should not be interpreted as necessarily representing the official policies or endorsements, either expressed or implied,
of IARPA, DOI/IBC, or the U.S. Government.

{
    \small
    \bibliographystyle{ieeenat_fullname}
    \bibliography{main}
}
 
\clearpage
\setcounter{page}{1}

\appendix

\section*{Appendix}

The appendix includes sections as follows:
\begin{itemize}
    \item \textbf{Section \ref{sec:A}}: Implementation details of both large-scale training and ablation studies.
    \item \textbf{Section \ref{sec:B}}: Flow evaluation.
    \item \textbf{Section \ref{sec:C}}: Additional quantitative results.
    \item \textbf{Section \ref{sec:D}}: Additional qualitative comparisons.
\end{itemize}

\section{Implementation Details}
\label{sec:A}

In this section, we provide implementation and training details
for all experiments of \ours{}.

\subsection{Large-scale Training Settings}
\paragraph{Full Model -- \ours.} Our primary model is obtained by fine-tuning the $\pi^3$~\cite{wang2025pi} backbone and adding a dense flow head. Given $N$ input images, a DINOv2~\cite{oquab2023dinov2} encoder extracts patch tokens which are processed by a $36$-layer multi-view transformer to produce per-image latent representations. These representations are then decoded by two task-specific transformer branches into geometry features and camera features. A point head predicts local pointmaps from the geometry features, while a camera head predicts camera poses from the camera features. In $\pi^3$, the predicted local pointmaps are supervised in a permutation-equivariant manner without selecting a reference view. For image $I_i$ in the $N$-view input, the network predicts a pixel-aligned pointmap $\hat{\mathbf{P}}_i\in\mathbb{R}^{H\times W\times 3}$ in its own camera coordinate system. Due to scale ambiguity, the predicted pointmaps across all views share a single global scale factor $s$. During training, an optimal scale $s^{*}$ is estimated from the ground-truth geometry and used to compute the reconstruction error:
\begin{equation}
\mathcal{L}_{\text{points}}
=\frac{1}{3NHW}\sum_{i=1}^{N}\sum_{u}\frac{1}{\mathbf{D}_i(u)}\|s^{*}\hat{\mathbf{P}}_i(u)-\mathbf{P}_i(u)\|_{1},
\end{equation}
where $\hat{\mathbf{P}}_i(u)$ and $\mathbf{P}_i(u)$ denote the predicted and ground-truth 3D points at pixel $u$, and $\mathbf{D}_i(u)$ is its ground-truth depth. Camera supervision in $\pi^3$ is defined on relative poses.
Given the predicted camera poses $\hat{\mathbf{T}}_i$ and $\hat{\mathbf{T}}_j$, the relative pose from view $j$ to $i$ is defined as $\hat{\mathbf{T}}_{i\leftarrow j}=\hat{\mathbf{T}}_i^{-1}\hat{\mathbf{T}}_j$. The camera loss averages over all ordered pairs and is a weighted sum of a rotation term and a translation term :
\begin{equation}
\mathcal{L}_{\text{cam}}=\frac{1}{N(N-1)}\sum_{i\neq j}\Big(\mathcal{L}_{\text{rot}}(i,j)+\lambda_{\text{trans}}\mathcal{L}_{\text{trans}}(i,j)\Big),
\end{equation}
where $\mathcal{L}_{\text{rot}}(i,j)$ is the geodesic distance between relative rotations,
\begin{equation}
\mathcal{L}_{\text{rot}}(i,j)=\arccos\!\left(\frac{\operatorname{Tr}\!\big(\mathbf{R}_{i\leftarrow j}^{\top}\hat{\mathbf{R}}_{i\leftarrow j}\big)-1}{2}\right),
\label{equ: relative_rot}
\end{equation}
and $\mathcal{L}_{\text{trans}}(i,j)$ is a Huber loss on relative translation after resolving scale ambiguity using the same $s^*$ computed from pointmap alignment. We refer the reader to $\pi^3$~\cite{wang2025pi} for further details. Building on this backbone and objectives, \ours{} introduces an additional dense factored flow supervision term via the added flow head, while keeping the original $\pi^3$ geometry and camera supervision unchanged. 

\paragraph{Full Model -- \oursvggt.} 
We also instantiate a model variant by fine-tuning the VGGT~\cite{wang2025vggt} backbone. It similarly employs a DINOv2~\cite{oquab2023dinov2} encoder, followed by a 48-layer multi-view transformer with alternating frame-wise and global self-attention blocks that produces camera features and geometry features. These features are then processed by task-specific heads to predict camera poses and scene geometry.

Unlike $\pi^3$~\cite{wang2025pi}, VGGT predicts geometry and poses in a global frame -- the coordinate frame of the first image in the input. Following the insight of $\pi^3$, we supervise the camera and geometry using relative quantities to remove the dependence on an absolute frame. We first normalize the predicted 3D pointmaps using their mean distance to the centroid, and this value serves as the scene scale implicitly learned by the model. For camera supervision, we adopt the same relative rotation loss $\mathcal{L}_{\text{rot}}$ as $\pi^3$ (Eq.~\ref{equ: relative_rot}), but replace their relative translation loss with a center loss $\mathcal{L}_{\text{center}}$ directly on the predicted camera centers after computing an optimal alignment, which we empirically find to perform better. Specifically, for the predicted camera centers $\{\hat{\mathbf{c}}_i\}$, we compute an optimal rigid transform $(\mathbf{R}^\ast, \mathbf{t}^\ast)$ 
that minimizes $\sum_i \|\mathbf{R}^\ast \hat{\mathbf{c}}_i + \mathbf{t}^\ast - \mathbf{c}_i\|_2^2$,
and aligns the predicted centers as $\hat{\mathbf{c}}_i^{\,\text{aligned}} = \mathbf{R}^\ast \hat{\mathbf{c}}_i + \mathbf{t}^\ast$.  
The camera center loss is defined as:
\begin{equation*}
\mathcal{L}_{\text{center}} =
\frac{1}{N} \sum_i
\big\|
\mathbf{c}_i^{\,\text{aligned}} - \mathbf{c}_i
\big\|_1,
\end{equation*}
where $N$ denotes the number of views in the input. 
For supervising the predicted global pointmaps, we introduce a permutation-equivariant training objective -- we estimate the best rigid transform that aligns the predicted pointmap $\hat{\mathbf{P}}_i$ to the ground truth $\mathbf{P}_i$ and computes an $\ell_2$ loss on the aligned points:
\begin{align*}
\mathcal{L}_{\text{point}}
&= \frac{1}{N}\min_{\mathbf{R}^\ast\in SO(3),\,\mathbf{t}^\ast\in\mathbb{R}^3}\sum_i
\|\mathbf{R}^\ast\hat{\mathbf{P}}_i + \mathbf{t}^\ast - \mathbf{P}_i\|_2.
\end{align*} 
To prevent coordinate drift over training, we regularize the mean position of all predicted points to stay centered around the origin:
\begin{equation}
\mathcal{L}_{\text{reg}} =
\left\|
\frac{1}{N}\sum_i \hat{\mathbf{P}}_i
\right\|_2.
\end{equation}

The depth head is supervised using the confidence-weighted regression loss introduced in DUSt3R~\cite{wang2023DUSt3R}. Specifically, for each pixel $u$, we regress the predicted depth $\hat{\textbf{D}}_i[u]$ to the ground-truth depth $\textbf{D}_i[u]$ using the predicted uncertainty $\mathbf{\Sigma}_i[u]$ as a per-pixel confidence weight. The depth loss is:

\begin{equation}
\mathcal{L}_{\text{depth}}
= \frac{1}{N} \sum_{i=1}^{N}
\left(
\left\|\mathbf{\Sigma}_i \odot (\hat{\mathbf{D}}_i - \mathbf{D}_i)\right\|_2
\;-\;
\sum_{u} \alpha \log \mathbf{\Sigma}_i[u]
\right),
\end{equation}

The total loss for supervising camera and geometric predictions is therefore:
\begin{equation}
\mathcal{L}_{\text{supervised}} = 
\mathcal{L}_{\text{rot}} + 
\mathcal{L}_{\text{center}} + 
\mathcal{L}_{\text{depth}} + 
\mathcal{L}_{\text{point}}+
\beta \mathcal{L}_{\text{reg}}.
\end{equation}

\paragraph{Flow Head.} 
We predict dense flow between a pair of images by fusing source-view geometry features with target-view camera features and decoding the fused representation using a DPT head~\cite{ranftl2021vision}. For \ours{} ($\pi^3$ backbone), the geometry features describe local structure within each view. Consequently, in addition to combining the source-view geometry features with target-view camera features, we also incorporate source-view camera features to transform local point representations into a consistent global geometry. Specifically, we first average the target-view camera features to obtain a global camera token, concatenate it with the source-view geometry features and source-view camera features, and fuse them using a lightweight MLP. We aggregate features from multiple transformer layers and decode the fused representation with a DPT head to predict dense flow. For \oursvggt{} (VGGT backbone), the predicted geometric elements are defined directly in a global coordinate frame. We therefore concatenate only the source-view geometry features with the target-view camera features and decode them using the same DPT head to produce dense flow.

\paragraph{Training Procedure.} We adopt the dynamic batch sizing strategy of $\pi^3$. Each GPU processes 24 images, with the batch size varying from 2 to 24 images. Image resolutions are randomly sampled such that the total pixel count lies between 100k and 255k, and the aspect ratio is drawn from $[0.5, 2.0]$. We train the models in multiple stages. For the VGGT backbone, we first fine-tune the model on labeled 3D data without flow supervision to adapt it to the permutation-equivariant formulation described above. For both \ours{} and \oursvggt{}, we initialize the model from the pretrained backbone and train only the factored flow head for 50k steps while freezing the backbone, using 8 H100 GPUs with a learning rate of $5\times10^{-5}$. After this warm-up stage, we unfreeze the entire model and train end-to-end for another 100k steps on both labeled and unlabeled data, using 8 H100 GPUs with a learning rate of $2\times10^{-5}$. This two-stage procedure stabilizes geometry learning under full supervision before introducing large-scale flow supervision from unlabeled videos.

\subsection{Ablation Studies} For the ablation studies in Sec.~\ref{sec:abla1}, we use a compact model that preserves the overall architecture of \oursvggt{} but with reduced capacity for efficiency. Concretely, we adopt a 224$\times$224 DINOv2 backbone and a 24-layer multi-view transformer, while keeping the prediction heads and training objectives unchanged. Despite being significantly smaller, this model remains architecturally compatible with the full variant and supports all ablation analyses. All ablation models are trained from scratch for 160k iterations on 4 H100 GPUs. We use Adam with a learning rate of $2\times10^{-5}$.

\subsection{Evaluating Dynamic Scene Reconstruction}
For dynamic datasets such as Kinetics-700 and EPIC-KITCHENS, which lack annotated geometry and camera poses, we generate pseudo labels using MegaSAM~\cite{li2025megasam}. For Kinetics-700, evaluation is performed on a curated subset of 60 sequences. For EPIC-KITCHENS, we process 100 frames per sequence to obtain pseudo annotations.

To fairly assess temporal reasoning under varying motion magnitudes, we adopt a stride-based sampling protocol. For each dataset, we predefine a set of temporal strides, and at evaluation time randomly sample one stride per sequence. The input consists of 8 ordered frames selected according to the sampled stride. While the frames preserve their original temporal order, they are not necessarily temporally adjacent. As a result, smaller strides emphasize short-term motion handling, whereas larger strides challenge the model’s ability to maintain long-range dynamic consistency.

\section{Flow Evaluation}
\label{sec:B}
While our method leverages flow supervision, accurate dense correspondence prediction is not its primary objective. Instead, flow serves as an intermediate supervisory signal to improve visual geometry learning. Owing to the factored formulation, correspondences are predicted by combining geometry and camera representations, rather than by directly matching dense features across views as in tracking-based flow heads. Consequently, the model does not explicitly optimize for pixel-level matching accuracy.

Tab.~\ref{tab:dataset_loss_results} reports standard flow metrics (AEPE and EPE@5px) on SpatialVID. We observe that the factored formulation achieves lower flow accuracy than tracking-based predictors, yet consistently yields better camera pose estimation and geometry reconstruction. This indicates that the benefit of flow supervision does not stem from precise correspondence prediction per se, but from the geometric constraints it imposes during training.

\begin{table}[H]
\centering
\small
\resizebox{\linewidth}{!}{
\begin{tabular}{ll cccc>{\columncolor{lightblue}}c >{\columncolor{lightblue}}c }
\toprule
\textbf{Model Variant} & \textbf{Data} \textbf{(\# Seqs)}  & \textbf{RRA}$\uparrow$ & \textbf{RTA}$\uparrow$& \textbf{CD} $\downarrow$ & \textbf{MSE} $\downarrow$ & \textbf{AEPE} $\downarrow$ & \textbf{EPE@5px} $\uparrow$\\
\midrule
\expb{}  & \textbf{+} SpatialVID ($3$K) & 61.23 & 56.12 &0.158&0.710 & 30.24&24.15 \\
\expc{}  &  \textbf{+} SpatialVID ($3$K) & 68.56 & 62.95&0.107&0.628 &\textbf{18.82}&\textbf{43.26}\\
\expd{} &  \textbf{+} SpatialVID ($3$K) & 76.26 & 68.84 & 0.103 & 0.598&23.29&36.42\\
\bottomrule
\end{tabular}
}
\caption{\textbf{Pose, geometry, and flow evaluation.}}
\label{tab:dataset_loss_results}
\end{table}

\section{Additional Quantitative Results}
\label{sec:C}

Beyond the main results, we provide additional quantitative evaluations on several more datasets (see Tables~\ref{tab:Kinetics700}, \ref{tab:epickitchens}, 
\ref{tab:sintel}, \ref{tab:bonn}, \ref{tab:7scenes}, 
\ref{tab:nrgbd}, \ref{tab:scannet}, \ref{tab:co3d}).
These include the datasets from the main text, for which we report all relevant metrics, as well as several representative static-scene benchmarks such as 7-Scenes~\cite{shotton2013scene}, ScanNet~\cite{dai2017scannet}, CO3Dv2~\cite{reizenstein2021common}, and NRGBD~\cite{azinovic2022neural}. 
For camera evaluation, we follow the standard protocols for each setting: static-scene datasets are evaluated using Relative Rotation Accuracy (RRA), Relative Translation Accuracy (RTA), and the Area Under the Curve which combines the first two metrics, while dynamic-scene datasets are evaluated using Absolute Trajectory Error (ATE), Relative Pose Error for translation (RPE trans), and Relative Pose Error for rotation (RPE rot).
For geometry evaluation, we adopt the same metrics for both static and dynamic datasets, reporting Accuracy, Completeness, Chamfer Distance, Mean-Squared Error, and F-score at 2\% and 5\% thresholds. Across these results, we observe that \ours{} shows a general improvement over the baselines, including methods trained with substantially more labeled data (\eg, the non-public synthetic data for training $\pi^3$~\cite{wang2025pi}).

\paragraph{Efficacy of Pseudo-labeled 3D Supervision.}
A natural alternative for leveraging unlabeled video data is to directly generate pseudo ground-truth 3D geometry and camera poses using off-the-shelf optimization-based systems such as MegaSAM~\cite{li2025megasam}. To examine this possibility, we conduct an ablation study in which we generate pseudo 3D and pose labels for 3K SpatialVID sequences using MegaSAM. As shown in Tab.~\ref{tab:supp_ablation_scaling}, training with these MegaSAM-generated pseudo-labels -- together with 1K labeled OmniWorld sequences -- performs worse than our proposed factored flow supervision. This performance gap suggests that directly regressing 3D geometry and poses from optimization-based `teachers' may introduce substantial noise and artifacts. In contrast, our formulation provides a cleaner and more effective supervisory signal for leveraging unlabeled video data. Moreover, generating pseudo-labels with MegaSAM is computationally expensive and difficult to scale. In comparison, our flow-based supervision is not only more effective but also easier to deploy across diverse unlabeled datasets.

\begin{table}[H]
\centering
\small
\resizebox{\linewidth}{!}{
\begin{tabular}{llcccc}
\toprule
\multicolumn{2}{c}{\textbf{Data}} & \multicolumn{4}{c}{\textbf{Metrics}} \\
\cmidrule(lr){1-2}\cmidrule(lr){3-6}
\textbf{Labeled (\# Seqs)} & \textbf{Unlabeled (\# Seqs)} & \textbf{RRA}$\uparrow$ & \textbf{RTA}$\uparrow$ & \textbf{CD}$\downarrow$ & \textbf{MSE}$\downarrow$ \\
\midrule
OmniWorld ($1$K) & \multicolumn{1}{c}{-} & 66.01 & 62.37 &0.105&0.637\\
\midrule
OmniWorld ($1$K) &  SpatialVID ($3$K) & 76.26 & 68.84 & 0.103 & 0.598\\
\rowcolor{lightblue}
OmniWorld ($1$K)+SpatialVID ($3$K) &   \multicolumn{1}{c}{-} & 75.99 & 65.30& 0.108&0.625\\
\midrule
OmniWorld ($4$K)  &  \multicolumn{1}{c}{-} & 78.68 & 70.26 & 0.080&0.565 \\
\bottomrule
\end{tabular}}
\caption{\textbf{Ablating pseudo-labeled 3D as supervision.}}
\label{tab:supp_ablation_scaling}
\end{table}

\section{Additional Qualitative Comparisons}
\label{sec:D}
We evaluate our model on a diverse collection of in-the-wild videos spanning both static and dynamic scenes, including everyday scenarios with people, animals, vehicles, and complex background clutter. As shown in Fig.~\ref{fig:supp_vis_1} and Fig.~\ref{fig:supp_vis_2}, across these varied examples, our method consistently produces competitive or superior 3D geometry compared to baseline methods. We additionally provide full point-cloud visualizations in video format on the project website.

\setlength{\tabcolsep}{4pt} 
\begin{table*}[t]
\centering
\footnotesize
\resizebox{0.9\linewidth}{!}{%
\begin{tabular}{l ccccccccc}
\toprule
\multirow{2}{*}{\textbf{Methods}} &
\multicolumn{9}{c}{\textbf{Kinetics700}}\\
\cmidrule(lr){2-10} 
 & ATE$\downarrow$ & RPE trans$\downarrow$ & RPE rot$\downarrow$
 & Acc.$\downarrow$ & Comp.$\downarrow$ & CD$\downarrow$ & MSE$\downarrow$ & f-score@2\%$\uparrow$ & f-score@5\%$\uparrow$  \\
\midrule
DUSt3R & 0.045 & 0.063 & 9.343 & 0.083 & \third{0.106} & 0.095 & 0.366 & 0.317 & 0.533 \\
CUT3R & \third{0.019} & \third{0.027} & 1.988 & \third{0.070} & \best{0.076} & \best{0.073} & \third{0.303} & \second{0.352} & \third{0.573} \\
VGGT & 0.027 & 0.038 & \third{1.392} & 0.088 & 0.120 & 0.104 & 0.347 & 0.258 & 0.479 \\
$\pi^{3}$ & \second{0.016} & \second{0.023} & \second{1.006} & \best{0.059} & 0.097 & \third{0.078} & \second{0.267} & \third{0.347} & \second{0.585} \\
Flow3r & \best{0.013} & \best{0.018} & \best{0.830} & \second{0.062} & \second{0.092} & \second{0.077} & \best{0.256} & \best{0.371} & \best{0.599} \\
\bottomrule
\end{tabular}
}
\caption{\textbf{Comparison on Kinetics700.}}
\label{tab:Kinetics700}
\end{table*}

\vspace{-8pt}

\setlength{\tabcolsep}{4pt} 
\begin{table*}[t]
\centering
\footnotesize
\resizebox{0.9\linewidth}{!}{%
\begin{tabular}{l ccccccccc}
\toprule
\multirow{2}{*}{\textbf{Methods}} &
\multicolumn{9}{c}{\textbf{EPIC-KITCHENS}}\\
\cmidrule(lr){2-10} 
 & ATE$\downarrow$ 
 & RPE trans$\downarrow$ 
 & RPE rot$\downarrow$
 & Acc.$\downarrow$ 
 & Comp.$\downarrow$ 
 & CD$\downarrow$ 
 & MSE$\downarrow$ 
 & f-score@2\%$\uparrow$ 
 & f-score@5\%$\uparrow$  \\
\midrule
DUSt3R 
& 0.077 & 0.110 & 8.492
& 0.100 & 0.092 & 0.096 & 0.312 & \third{0.385} & 0.528 \\

CUT3R 
& 0.056 & 0.081 & 4.709
& 0.085 & 0.095 & 0.090 & 0.338 & 0.297 & 0.493 \\
VGGT 
& \best{0.032} & \second{0.049} & \second{3.025} 
& \best{0.061} & \third{0.069} & \third{0.065} & \third{0.220} & \second{0.415} & \third{0.617} \\
$\pi^{3}$ 
& \best{0.032} & \second{0.043} & \second{3.025}
& \third{0.069} & \second{0.058} & \best{0.059} & \second{0.200} & \best{0.459} & \second{0.620} \\
Flow3r
& \best{0.032} & \best{0.037} & \best{2.729}
& \second{0.066} & \best{0.056} & \best{0.059} & \best{0.199} & 0.339 & \best{0.622} \\

\bottomrule
\end{tabular}
}
\caption{\textbf{Comparison on EPIC-KITCHENS.} }
\label{tab:epickitchens}
\end{table*}

\setlength{\tabcolsep}{4pt} 
\begin{table*}[t]
\centering
\footnotesize
\resizebox{0.9\linewidth}{!}{%
\begin{tabular}{l ccccccccc}
\toprule
\multirow{2}{*}{\textbf{Methods}} &
\multicolumn{9}{c}{\textbf{Sintel}}\\
\cmidrule(lr){2-10} 
 & ATE$\downarrow$ & RPE trans$\downarrow$ & RPE rot$\downarrow$
 & Acc.$\downarrow$ & Comp.$\downarrow$ & CD$\downarrow$ & MSE$\downarrow$ & f-score@2\%$\uparrow$ & f-score@5\%$\uparrow$  \\
\midrule
DUSt3R & 0.152 & 0.179 & 15.166 & 0.255 & 0.224 & 0.240 & 0.622 & \third{0.124} & 0.271 \\
CUT3R & 0.111 & 0.128 & 1.998 & 0.221 & 0.269 & 0.245 & 0.676 & 0.111 & 0.217 \\
VGGT & \third{0.090} & \third{0.086} & \third{1.220} & \third{0.217} & \third{0.227} & \third{0.222} & \third{0.595} & 0.105 & \third{0.311} \\
$\pi^{3}$ & \second{0.060} & \second{0.066} & \second{1.122} & \second{0.189} & \second{0.191} & \second{0.190} & \second{0.523} & \second{0.141} & \second{0.317} \\
Flow3r & \best{0.048} & \best{0.058} & \best{0.920} & \best{0.153} & \best{0.104} & \best{0.129} & \best{0.426} & \best{0.201} & \best{0.404} \\
\bottomrule
\end{tabular}
}
\caption{\textbf{Comparison on Sintel.} }
\label{tab:sintel}
\end{table*}

\vspace{-8pt}

\setlength{\tabcolsep}{4pt} 
\begin{table*}[t]
\centering
\footnotesize
\resizebox{0.9\linewidth}{!}{%
\begin{tabular}{l ccccccccc}
\toprule
\multirow{2}{*}{\textbf{Methods}} &
\multicolumn{9}{c}{\textbf{Bonn}}\\
\cmidrule(lr){2-10} 
 & ATE$\downarrow$ & RPE trans$\downarrow$ & RPE rot$\downarrow$
 & Acc.$\downarrow$ & Comp.$\downarrow$ & CD$\downarrow$ & MSE$\downarrow$ & f-score@2\%$\uparrow$ & f-score@5\%$\uparrow$  \\
\midrule
DUSt3R & 0.055 & 0.113 & 6.384 & 0.029 & 0.037 & 0.033 & 0.116 & 0.546 & 0.800 \\
CUT3R & 0.021 & \second{0.095} & 6.349 & 0.018 & \second{0.023} & \third{0.021} & 0.088 & \third{0.658} & \third{0.899} \\
VGGT & \second{0.011} & 0.096 & \best{6.021} & \third{0.016} & 0.028 & \third{0.021} & \third{0.082} & 0.644 & 0.884 \\
$\pi^{3}$ & \third{0.016} & \second{0.095} & \second{6.240} & \second{0.014} & \third{0.025} & \second{0.019} & \second{0.076} & \second{0.669} & \second{0.905} \\
Flow3r & \best{0.009} & \best{0.094} & \third{6.262} & \best{0.009} & \best{0.018} & \best{0.013} & \best{0.052} & \best{0.801} & \best{0.954} \\
\bottomrule
\end{tabular}
}
\caption{\textbf{Comparison on Bonn.}}
\label{tab:bonn}
\end{table*}

\setlength{\tabcolsep}{4pt} 
\begin{table*}[!h]
\centering
\footnotesize
\resizebox{0.9\linewidth}{!}{%
\begin{tabular}{l ccccccccc}
\toprule
\multirow{2}{*}{\textbf{Methods}} &
\multicolumn{9}{c}{\textbf{7-scenes}}\\
\cmidrule(lr){2-10} 
 & RRA@30$\uparrow$ & RTA@30$\uparrow$ & AUC@30$\uparrow$
 & Acc.$\downarrow$ & Comp.$\downarrow$ & CD$\downarrow$ & MSE$\downarrow$ & f-score@2\%$\uparrow$ & f-score@5\%$\uparrow$  \\
\midrule
DUSt3R & 100.0 & 76.59 & 55.07 & \third{0.047} & 0.053 & \third{0.050} & \third{0.170} & \third{0.457} & \third{0.714} \\
CUT3R &  100.0 & 81.15 & 59.72 & 0.055 & \third{0.049} & 0.052 & 0.183 & \third{0.457} & 0.695 \\
VGGT & 100.0 & \third{86.51} & \third{69.77} & 0.052 & 0.056 & 0.054 & 0.182 & 0.395 & 0.665 \\
$\pi^{3}$ & 100.0 & \second{87.69} & \second{71.80} & \second{0.039} & \second{0.045} & \second{0.042} & \second{0.169} & \second{0.473} & \second{0.737} \\
Flow3r & 100.0 & \best{91.66} & \best{75.76} & \best{0.025} & \best{0.016} & \best{0.020} & \best{0.102} & \best{0.729} & \best{0.807} \\
\bottomrule
\end{tabular}
}
\caption{\textbf{Comparison on 7-scenes.} }
\label{tab:7scenes}
\end{table*}

\vspace{-8pt}

\setlength{\tabcolsep}{4pt} 
\begin{table*}[!htb]
\centering
\footnotesize
\resizebox{0.9\linewidth}{!}{%
\begin{tabular}{l ccccccccc}
\toprule
\multirow{2}{*}{\textbf{Methods}} &
\multicolumn{9}{c}{\textbf{NRGBD}}\\
\cmidrule(lr){2-10} 
 & RRA@30$\uparrow$ & RTA@30$\uparrow$ & AUC@30$\uparrow$
 & Acc.$\downarrow$ & Comp.$\downarrow$ & CD$\downarrow$ & MSE$\downarrow$ & f-score@2\%$\uparrow$ & f-score@5\%$\uparrow$  \\
\midrule
DUSt3R & 100.0 & 93.25 & 76.04 & 0.031 & 0.024 & 0.027 & 0.063 & 0.662 & 0.865 \\
CUT3R & 100.0 & 95.63 & 76.90 & 0.042 & 0.019 & 0.030 & 0.075 & 0.549 & 0.808  \\
VGGT & 100.0 & \second{99.21} & \second{93.13} & \third{0.015} & \third{0.010} & \third{0.012} & \third{0.032} & \third{0.833} & \third{0.964} \\
$\pi^{3}$ & 100.0 & \third{99.20} & \third{90.40} & \second{0.010} & \second{0.007} & \second{0.008} & \second{0.021} & \second{0.910} & \second{0.983} \\
Flow3r & 100.0 & \best{99.60} & \best{94.40} & \best{0.008} & \best{0.005} & \best{0.007} & \best{0.018} & \best{0.938} & \best{0.992} \\
\bottomrule
\end{tabular}
}
\caption{\textbf{Comparison on NRGBD.} }
\label{tab:nrgbd}
\end{table*}

\vspace{-12pt}
\setlength{\tabcolsep}{4pt} 
\begin{table*}[!htb]
\centering
\footnotesize
\resizebox{0.9\linewidth}{!}{%
\begin{tabular}{l ccccccccc}
\toprule
\multirow{2}{*}{\textbf{Methods}} &
\multicolumn{9}{c}{\textbf{Scannet}}\\
\cmidrule(lr){2-10} 
 & RRA@30$\uparrow$ & RTA@30$\uparrow$ & AUC@30$\uparrow$
 & Acc.$\downarrow$ & Comp.$\downarrow$ & CD$\downarrow$ & MSE$\downarrow$ & f-score@2\%$\uparrow$ & f-score@5\%$\uparrow$  \\
\midrule
DUSt3R & \best{100.0} & 57.14 & 31.56 & 0.031 & 0.032 & 0.032 & 0.092 & 0.591 & 0.831 \\
CUT3R  & 99.39 & 71.39 & 42.30 & 0.053 & 0.034 & 0.043 & 0.132 & 0.499 & 0.740 \\
VGGT & \best{100.0} & \best{93.71} & \best{71.37} & \best{0.015} & \second{0.019} & \second{0.017} & \best{0.053} & \second{0.769} & \second{0.931} \\
$\pi^{3}$ & 99.75 & \third{91.14} & \third{69.39} & \third{0.016} & \third{0.022} & \third{0.019} & \third{0.058} & \third{0.762} & \third{0.930} \\
Flow3r & \best{100.0} & \second{92.89} & \second{71.00} & \best{0.015} & \best{0.018} & \best{0.016} & \best{0.053} & \best{0.780} & \best{0.943} \\
\bottomrule
\end{tabular}
}
\caption{\textbf{Comparison on Scannet.}}
\label{tab:scannet}
\end{table*}

\vspace{-12pt}

\setlength{\tabcolsep}{4pt} 
\begin{table*}[t]
\centering
\footnotesize
\resizebox{0.9\linewidth}{!}{%
\begin{tabular}{l ccccccccc}
\toprule
\multirow{2}{*}{\textbf{Methods}} &
\multicolumn{9}{c}{\textbf{Co3Dv2}}\\
\cmidrule(lr){2-10} 
 & RRA@30$\uparrow$ & RTA@30$\uparrow$ & AUC@30$\uparrow$
 & Acc.$\downarrow$ & Comp.$\downarrow$ & CD$\downarrow$ & MSE$\downarrow$ & f-score@2\%$\uparrow$ & f-score@5\%$\uparrow$  \\
\midrule
DUSt3R & 97.07 & 90.91 & 66.83 & 0.036 & 0.069 & \best{0.033} & 0.223 & 0.567 & 0.783 \\
CUT3R & 93.85 & 90.68 & 68.11 & 0.047 & 0.082 & 0.064 & 0.278 & 0.507 & 0.737  \\
VGGT & \third{98.41} & \third{97.27} & \third{87.62} & \second{0.022} & \best{0.051} & \second{0.036} & \second{0.151} & \second{0.707} & \second{0.874} \\
$\pi^{3}$ & \second{98.82} & \second{97.49} & \best{90.53} & \second{0.022} & \second{0.052} & \third{0.037} & \second{0.151} & \second{0.707} & \second{0.874} \\
Flow3r & \best{98.84} & \best{97.62} & \second{90.41} & \best{0.020} & \second{0.052} & \third{0.037} & \best{0.150} & \best{0.709} & \best{0.876} \\
\bottomrule
\end{tabular}
}
\caption{\textbf{Comparison on Co3Dv2.}}
\label{tab:co3d}
\end{table*}
\vspace{-12pt}

\begin{figure*}[t]
  \centering
  \includegraphics[width=0.9\linewidth]{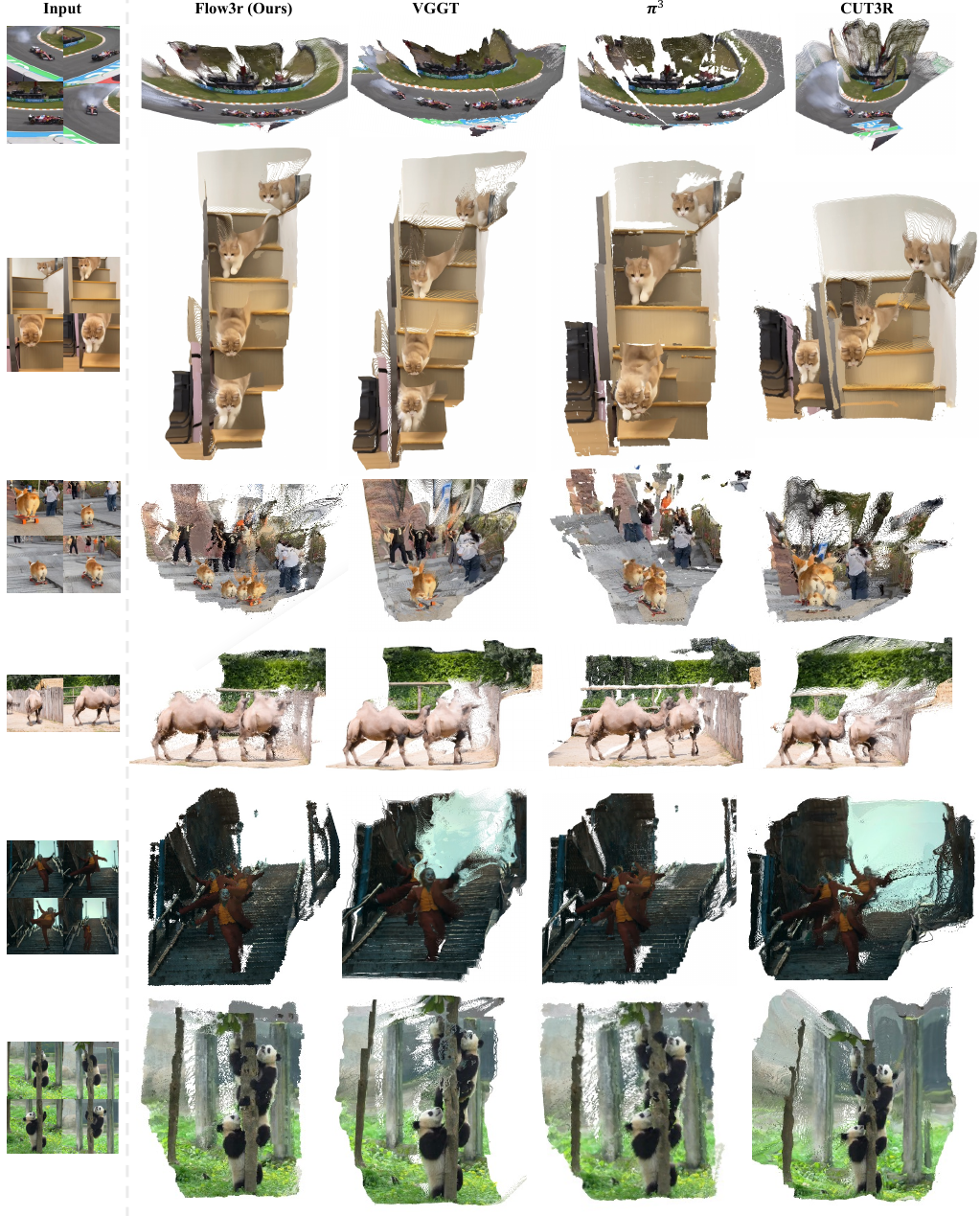}
  \caption{\textbf{More qualitative results on dynamic scenes.} \ours{} effectively handles dynamic components, yielding cleaner geometry than prior feed-forward methods.}
  \label{fig:supp_vis_1}
\end{figure*}

\begin{figure*}[!t]
  \centering
  \includegraphics[width=0.9\linewidth]{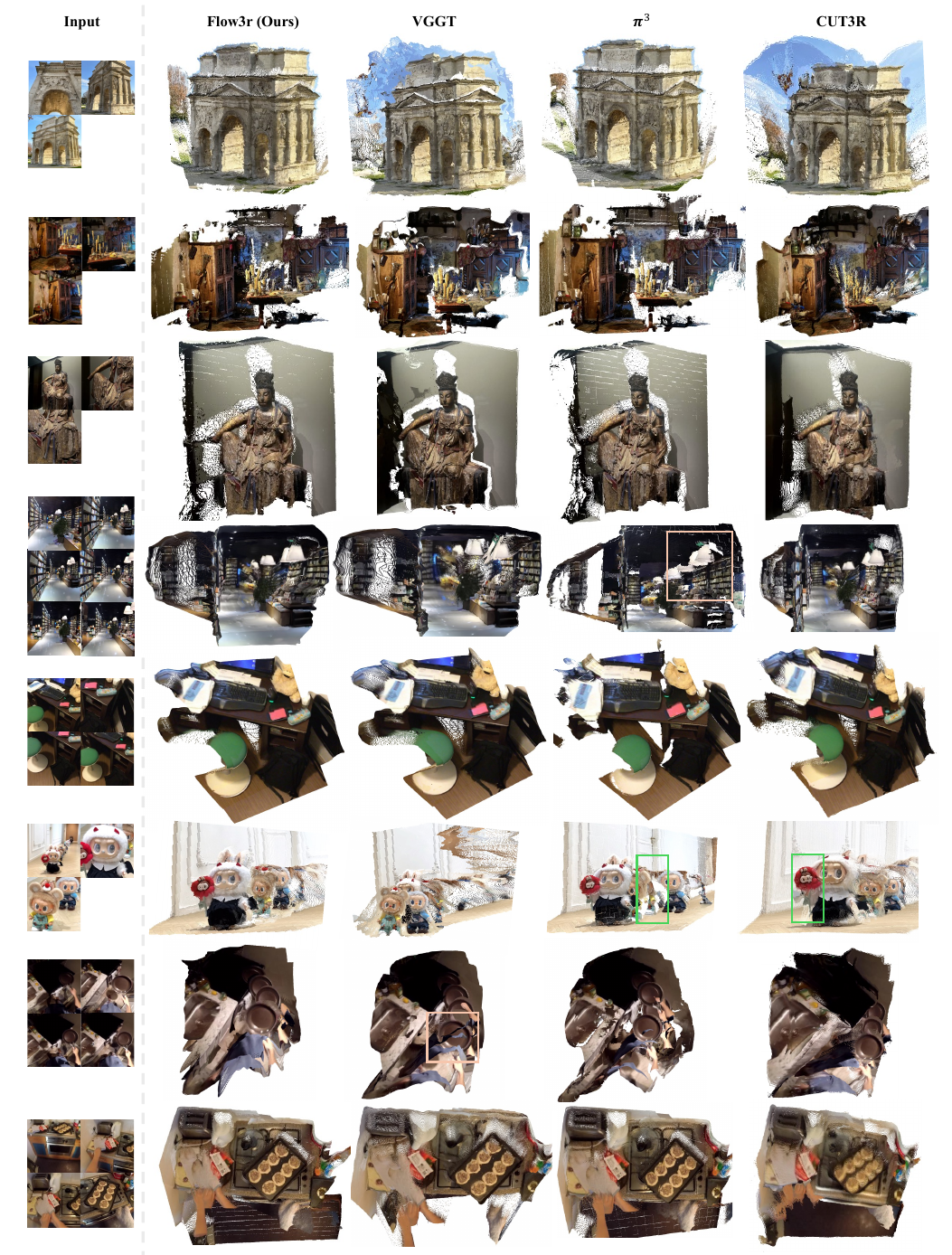}
    \caption{\textbf{More qualitative results on static scenes and interaction videos.} The first six examples show static scenes and the last two examples feature interaction videos. \ours{} tends to produce more stable geometry under motion, though some challenging cases still show noticeable artifacts.}
  \label{fig:supp_vis_2}
\end{figure*}

\end{document}